# Homomorphisms between fuzzy information systems revisited

Ping Zhu[a,b,*], Qiaoyan Wen[b]

[a]*School of Science, Beijing University of Posts and Telecommunications, Beijing 100876, China*
[b]*State Key Laboratory of Networking and Switching, Beijing University of Posts and Telecommunications, Beijing 100876, China*

**Abstract**

Recently, Wang et al. discussed the properties of fuzzy information systems under homomorphisms in the paper [C. Wang, D. Chen, L. Zhu, Homomorphisms between fuzzy information systems, Applied Mathematics Letters 22 (2009) 1045-1050], where homomorphisms are based upon the concepts of consistent functions and fuzzy relation mappings. In this paper, we classify consistent functions as predecessor-consistent and successor-consistent, and then proceed to present more properties of consistent functions. In addition, we improve some characterizations of fuzzy relation mappings provided by Wang et al.

## 1. Introduction

Information systems [5, 8], also called knowledge representation systems, are a formalism for representing knowledge about some objects in terms of attributes (e.g., color) and values of attributes (e.g., green). Over the last decades, the concept of information systems has gained considerable attention, including some successful applications in information processing, decision, and control (see, for example, [1, 7, 10, 11, 16, 18, 20]). To study transformations of information systems while preserving their basic functions, a mathematical tool, homomorphism, has been introduced and investigated in the literature [2, 3, 6, 14, 15, 19, 21].

Most recently, Wang et al. discussed the properties of fuzzy information systems under homomorphisms in [12, 13]. In particular, they showed that attribute reductions in the original fuzzy information system and homomorphic image are equivalent to each other under a homomorphism. Thereby, homomorphisms are applicable in simulation of big systems by their smaller homomorphic images. The concept of homomorphisms, in turn, is based upon the notions of consistent functions and fuzzy relation mappings. Some basic properties of consistent functions and fuzzy relation mappings have been presented in [12].

In this paper, we revisit the homomorphisms between fuzzy information systems. More concretely, we classify consistent functions in [12] as predecessor-consistent and successor-consistent, and then proceed to present more properties of consistent functions. We improve some characterizations of fuzzy relation mappings provided in [12]. In particular, we present a new relationship between fuzzy neighborhoods and fuzzy relation mappings, which provides an approach to computing the fuzzy predecessor and fuzzy successor neighborhoods of an element of codomain with respect to the induced fuzzy relation. The theory presented here is helpful in establishing homomorphisms from the original fuzzy information system to a simpler fuzzy information system, which preserves some functions of the original system.

The remainder of this paper is structured as follows. In Section 2, we introduce predecessor-consistent and successor-consistent functions, and show that they are together equivalent to the concept of consistent functions in the sense of [12]. Some properties of predecessor-consistent and successor-consistent functions are also explored in this section. Based on the classification of consistent functions, we extend some characterizations of fuzzy relation mappings in Section 3 and conclude the paper in Section 4.

[*]Corresponding author
  *Email addresses:* `pzhubupt@gmail.com` (Ping Zhu), `wqy@bupt.edu.cn` (Qiaoyan Wen)

## 2. Consistent functions

For subsequent need, let us first review some notions on fuzzy set theory. For a detailed introduction to the notions, the reader may refer to [4, 9].

Let $U$ be a universal set. A *fuzzy set $A$*, or rather a *fuzzy subset $A$ of $U$*, is defined by a function assigning to each element $x$ of $U$ a value $A(x) \in [0, 1]$. We denote by $\mathcal{F}(U)$ the set of all fuzzy subsets of $U$. For any $A, B \in \mathcal{F}(U)$, we say that $A$ is contained in $B$ (or $B$ contains $A$), denoted by $A \subseteq B$, if $A(x) \leq B(x)$ for all $x \in U$, and we say that $A = B$ if and only if $A \subseteq B$ and $B \subseteq A$. The *support* of a fuzzy set $A$ is a crisp set defined as $\mathrm{supp}(A) = \{x \in X : A(x) > 0\}$. Whenever $\mathrm{supp}(A)$ is a finite set, say $\mathrm{supp}(A) = \{x_1, x_2, \ldots, x_n\}$, we may write $A$ in Zadeh's notation as

$$A = \frac{A(x_1)}{x_1} + \frac{A(x_2)}{x_2} + \cdots + \frac{A(x_n)}{x_n}.$$

For any family $\alpha_i$, $i \in I$, of elements of $[0, 1]$, we write $\vee_{i \in I} \alpha_i$ or $\vee\{\alpha_i : i \in I\}$ for the supremum of $\{\alpha_i : i \in I\}$, and $\wedge_{i \in I} \alpha_i$ or $\wedge\{\alpha_i : i \in I\}$ for the infimum. In particular, if $I$ is finite, then $\vee_{i \in I} \alpha_i$ and $\wedge_{i \in I} \alpha_i$ are the greatest element and the least element of $\{\alpha_i : i \in I\}$, respectively. Given $A, B \in \mathcal{F}(U)$, the *union* of $A$ and $B$, denoted $A \cup B$, is defined by $(A \cup B)(x) = A(x) \vee B(x)$ for all $x \in U$; the *intersection* of $A$ and $B$, denoted $A \cap B$, is given by $(A \cap B)(x) = A(x) \wedge B(x)$ for all $x \in U$.

For later need, let us recall Zadeh's extension principle. If $U$ and $V$ are two crisp sets and $f$ is a mapping from $U$ to $V$, then $f$ can be extended to a mapping from $\mathcal{F}(U)$ to $\mathcal{F}(V)$ in the following way: For any $A \in \mathcal{F}(U)$, $f(A) \in \mathcal{F}(V)$ is given by

$$f(A)(y) = \vee\{A(x) \mid x \in U \text{ and } f(x) = y\}$$

for all $y \in V$. Conversely, the mapping $f : U \longrightarrow V$ can induce a mapping $f^{-1}$ from $\mathcal{F}(V)$ to $\mathcal{F}(U)$ as follows: For any $B \in \mathcal{F}(V)$, $f^{-1}(B) \in \mathcal{F}(U)$ is defined by

$$f^{-1}(B)(x) = B(f(x))$$

for all $x \in U$.

Let $U$ be a finite and nonempty universal set, and suppose that $R \in \mathcal{F}(U \times U)$ is a fuzzy (binary) relation on $U$. For each $x \in U$, we associate it with a *fuzzy predecessor neighborhood $R_p^x$* and a *fuzzy successor neighborhood $R_s^x$* as follows:

$$\begin{array}{cccc} R_p^x : & U & \longrightarrow & [0, 1] \\ & y & \longmapsto & R(y, x) \end{array} \quad \text{and} \quad \begin{array}{cccc} R_s^x : & U & \longrightarrow & [0, 1] \\ & y & \longmapsto & R(x, y), \end{array}$$

that is, $R_p^x(y) = R(y, x)$ and $R_s^x(y) = R(x, y)$ for all $y \in U$. Clearly, for any $x \in U$, both the fuzzy predecessor neighborhood $R_p^x$ and the fuzzy successor neighborhood $R_s^x$ of $x$ are fuzzy subsets of $U$. Besides, more fuzzy neighborhoods can be defined; for example, one can define additional types of fuzzy neighborhoods of $x \in U$:

$$R_{p \wedge s}^x(y) = R(y, x) \wedge R(x, y) = R_p^x(y) \wedge R_s^x(y),$$
$$R_{p \vee s}^x(y) = R(y, x) \vee R(x, y) = R_p^x(y) \vee R_s^x(y).$$

Note that all the four fuzzy neighborhoods will reduce to usual neighborhoods in [17] if $R$ is a crisp binary relation (i.e., $R(x, y) \in \{0, 1\}$ for all $x, y \in U$).

With the concepts of fuzzy neighborhoods, we can introduce the following definition.

**Definition 2.1.** Let $U$ and $V$ be finite and nonempty universal sets, $R$ a fuzzy relation on $U$, and $f : U \longrightarrow V$ a mapping.

(1) The mapping $f$ is called a *predecessor-consistent function* with respect to $R$ if for any $x, y \in U$, $R_p^x = R_p^y$ whenever $f(x) = f(y)$.
(2) The mapping $f$ is called a *successor-consistent function* with respect to $R$ if for any $x, y \in U$, $R_s^x = R_s^y$ whenever $f(x) = f(y)$.

In other words, a mapping $f$ is predecessor-consistent (respectively, successor-consistent) if any two elements of $U$ with the same image under $f$ have the same fuzzy predecessor (respectively, fuzzy successor) neighborhood.

**Remark 2.1.** If $R$ is a crisp binary relation on $U$, then Definition 2.1 is exactly Definition 2.2 in [21]. In other words, Definition 2.1 is a generalization of Definition 2.2 in [21]. It should be noted that by Theorem 2.1 in [21], the concept of predecessor-consistent (respectively, successor-consistent) function is equivalent to that of type-1 (respectively, type-2) consistent function introduced in [15], when $R$ is a crisp binary relation.

To illustrate the definition, let us see a simple example.

**Example 2.1.** Set $U = \{x_1, x_2, \ldots, x_8\}$ and $V = \{y_1, y_2, \ldots, y_8\}$. Take

$$R = \frac{1}{(x_1, x_2)} + \frac{1}{(x_1, x_3)} + \frac{0.8}{(x_2, x_4)} + \frac{0.8}{(x_2, x_5)} + \frac{0.9}{(x_3, x_4)} + \frac{0.8}{(x_3, x_5)} +$$
$$\frac{0.7}{(x_4, x_6)} + \frac{0.7}{(x_4, x_7)} + \frac{0.7}{(x_5, x_6)} + \frac{0.7}{(x_5, x_7)} + \frac{0.9}{(x_6, x_8)} + \frac{0.9}{(x_7, x_8)}.$$

Define $f_k : U \longrightarrow V$, $k = 1, 2, 3$, as follows:

$$f_1(x_i) = \begin{cases} y_2 & \text{if } i = 2, 3, \\ y_i & \text{otherwise;} \end{cases} \quad f_2(x_i) = \begin{cases} y_4 & \text{if } i = 4, 5, \\ y_i & \text{otherwise;} \end{cases} \quad f_3(x_i) = \begin{cases} y_6 & \text{if } i = 6, 7, \\ y_i & \text{otherwise.} \end{cases}$$

Then by definition, it is easy to check that $f_1$ is predecessor-consistent (not successor-consistent) with respect to $R$, $f_2$ is successor-consistent (not predecessor-consistent) with respect to $R$, and $f_3$ is both predecessor-consistent and successor-consistent with respect to $R$.

Let us recall the concept of consistent function introduced in [12].

**Definition 2.2** ([12], Definition 2.2). Let $U$ and $V$ be finite universes, $R$ a fuzzy relation on $U$, and $f : U \longrightarrow V$ a mapping. Let

$$[x]_f = \{y \in U \mid f(y) = f(x)\}.$$

For any $x, y \in U$, if $R(x_1, y_1) = R(x_2, y_2)$ for any $(x_i, y_i) \in [x]_f \times [y]_f$, $i = 1, 2$, then $f$ is called a *consistent function* with respect to $R$.

As we will see, the consistent function in the sense of Definition 2.2 in [12] is nothing other than a function that is both predecessor-consistent and successor-consistent.

**Theorem 2.1.** *Let $U$ and $V$ be finite and nonempty universal sets, and $R$ a fuzzy relation on $U$. A mapping $f : U \longrightarrow V$ is consistent with respect to $R$ in the sense of Definition 2.2 if and only if it is both predecessor-consistent and successor-consistent with respect to $R$.*

*Proof.* We first prove the necessity. Suppose that $f : U \longrightarrow V$ is consistent with respect to $R$ in the sense of Definition 2.2. To see that $f$ is predecessor-consistent, letting $f(y_1) = f(y_2)$, we need to show that $R_p^{y_1} = R_p^{y_2}$, that is, $R_p^{y_1}(x) = R_p^{y_2}(x)$ for all $x \in U$. Since $f$ is consistent with respect to $R$ and $(x, y_i) \in [x]_f \times [y_1]_f$, $i = 1, 2$, we get by definition that $R(x, y_1) = R(x, y_2)$, which means that $R_p^{y_1}(x) = R_p^{y_2}(x)$. Therefore, $f$ is predecessor-consistent with respect to $R$. By the same token, we can show that $f$ is also successor-consistent with respect to $R$. Hence, the necessity holds.

Conversely, assume that $f$ is both predecessor-consistent and successor-consistent with respect to $R$. Let $x, y \in U$ and $(x_i, y_i) \in [x]_f \times [y]_f$, $i = 1, 2$. To show that $f$ is consistent, it suffices to verify that $R(x_1, y_1) = R(x_2, y_2)$. In fact, since $f$ is successor-consistent with respect to $R$ and $f(x_1) = f(x) = f(x_2)$, we see that $R_s^{x_1} = R_s^{x_2}$, which gives $R_s^{x_1}(y_1) = R_s^{x_2}(y_1)$, namely, $R(x_1, y_1) = R(x_2, y_1)$. On the other hand, because $f$ is predecessor-consistent with respect to $R$ and $f(y_1) = f(y) = f(y_2)$, we have that $R_p^{y_1} = R_p^{y_2}$, which yields $R_p^{y_1}(x_2) = R_p^{y_2}(x_2)$, namely, $R(x_2, y_1) = R(x_2, y_2)$. As a result, we obtain that $R(x_1, y_1) = R(x_2, y_1) = R(x_2, y_2)$, as desired. This completes the proof of the theorem. □

Recall that a fuzzy relation $R \in \mathcal{F}(U \times U)$ is called *reflexive* if $R(x, x) = 1$ for all $x \in U$; $R$ is said to be *symmetric* if $R(x, y) = R(y, x)$ for any $x, y \in U$; $R$ is called *transitive* or *max-min transitive* if $R(x, z) \geq R(x, y) \wedge R(y, z)$ for any $x, y, z \in U$. For a fuzzy relation $R$, the *inverse* $R^{-1}$ of $R$ is defined by

$$R^{-1}(x, y) = R(y, x)$$

for all $x, y \in U$. Clearly, $R$ is reflexive (respectively, transitive) if and only if $R^{-1}$ is reflexive (respectively, transitive), and $R$ is symmetric if and only if $R = R^{-1}$. Observe that the fuzzy predecessor neighborhood defined by $R$ is exactly the fuzzy successor neighborhood defined by $R^{-1}$, and conversely, the fuzzy successor neighborhood defined by $R$ is exactly the fuzzy predecessor neighborhood defined by $R^{-1}$. Formally, for each $x \in U$,

$$R_p^x(y) = R(y, x) = R^{-1}(x, y) = (R^{-1})_s^x(y), \tag{1}$$
$$R_s^x(y) = R(x, y) = R^{-1}(y, x) = (R^{-1})_p^x(y), \tag{2}$$

for all $y \in U$.

Let $R$ and $Q$ be two fuzzy relations on $U$. Defining $R \cup Q$ and $R \cap Q$ by fuzzy set-theoretic union and intersection, respectively, we have the following equations:

$$(R \cup Q)_p^x = R_p^x \cup Q_p^x, \tag{3}$$
$$(R \cup Q)_s^x = R_s^x \cup Q_s^x, \tag{4}$$
$$(R \cap Q)_p^x = R_p^x \cap Q_p^x, \tag{5}$$
$$(R \cap Q)_s^x = R_s^x \cap Q_s^x, \tag{6}$$

for any $x \in U$. They follow directly from the definitions of fuzzy predecessor and fuzzy successor neighborhoods.

The following proposition clarifies the relationship between predecessor-consistent functions and successor-consistent functions. As a result, we may think that predecessor-consistent functions and successor-consistent functions are symmetric in some sense.

**Proposition 2.1.** *Let $U$ and $V$ be finite and nonempty universal sets and $R$ a fuzzy relation on $U$.*

(1) *A mapping $f : U \longrightarrow V$ is predecessor-consistent with respect to $R$ if and only if it is successor-consistent with respect to $R^{-1}$.*
(2) *A mapping $f : U \longrightarrow V$ is successor-consistent with respect to $R$ if and only if it is predecessor-consistent with respect to $R^{-1}$.*

*Proof.* It follows immediately from Eqs. (1) and (2). □

If $R$ is a symmetric relation, then predecessor-consistent functions are exactly successor-consistent.

**Corollary 2.1.** *Let $U$ and $V$ be finite and nonempty universal sets. If the fuzzy relation $R$ on $U$ is symmetric, then a mapping $f : U \longrightarrow V$ is predecessor-consistent with respect to $R$ if and only if it is successor-consistent with respect to $R$.*

*Proof.* It follows immediately from Proposition 2.1 and the fact that $R^{-1} = R$ if $R$ is symmetric. □

In addition, a predecessor-consistent function is exactly successor-consistent when $R$ is reflexive and transitive. To prove this, it is handy with the following lemma.

**Lemma 2.1.** *Let $R$ be a reflexive and transitive fuzzy relation on $U$. Then for any $x, y \in U$, $R_p^x = R_p^y$ if and only if $R_s^x = R_s^y$.*

*Proof.* We only prove the necessity; the sufficiency can be verified in the same way. By contradiction, assume that $R_s^x \neq R_s^y$. Without loss of generality, suppose that there exists some $z \in U$ such that $R_s^x(z) > R_s^y(z)$. Then we see that $R(x, z) > R(y, z)$. Since $R$ is reflexive, we get that $R(y, x) = R_p^x(y) = R_p^y(y) = R(y, y) = 1$, namely, $R(y, x) = 1$. We thus have by the transitivity of $R$ that

$$R(y, z) \geq R(y, x) \wedge R(x, z) = R(x, z) > R(y, z),$$

namely, $R(y, z) > R(y, z)$, which is absurd. Consequently, $R_s^x = R_s^y$ and the necessity holds. □

The following theorem says that a mapping is predecessor-consistent if and only if it is successor-consistent, when the relation $R$ is reflexive and transitive.

**Theorem 2.2.** *Let U and V be finite and nonempty universal sets. If R is a reflexive and transitive fuzzy relation on U, then a mapping $f : U \longrightarrow V$ is predecessor-consistent with respect to R if and only if it is successor-consistent with respect to R.*

*Proof.* It is straightforward by Lemma 2.1. □

Recall that Eqs. (5) and (6) say that $(R \cap Q)_p^x = R_p^x \cap Q_p^x$ and $(R \cap Q)_s^x = R_s^x \cap Q_s^x$, respectively. Such equalities can be preserved under some mappings.

**Theorem 2.3.** *Let R and Q be fuzzy relations on U, and $f : U \longrightarrow V$ a mapping.*

(1) *If f is predecessor-consistent with respect to either R or Q, then $f((R \cap Q)_s^x) = f(R_s^x) \cap f(Q_s^x)$ for any $x \in U$.*
(2) *If f is successor-consistent with respect to either R or Q, then $f((R \cap Q)_p^x) = f(R_p^x) \cap f(Q_p^x)$ for any $x \in U$.*

*Proof.* (1) Without loss of generality, we may assume that $f$ is predecessor-consistent with respect to $R$. We first claim that if $z_1, z_2 \in U$ with $f(z_1) = f(z_2)$, then $R_s^x(z_1) = R_s^x(z_2)$ for any $x \in U$. In fact, since $f$ is predecessor-consistent with respect to $R$, we have by definition that $R_p^{z_1} = R_p^{z_2}$. This means that $R_p^{z_1}(x) = R_p^{z_2}(x)$, namely, $R(x, z_1) = R(x, z_2)$, for any $x \in U$. Hence, we get that $R_s^x(z_1) = R_s^x(z_2)$ for any $x \in U$. It follows from the claim that we may set $r_y = R_s^x(z)$ for any $z \in U$ with $f(z) = y$. To prove $f((R \cap Q)_s^x) = f(R_s^x) \cap f(Q_s^x)$, it is sufficient to show that $f(R_s^x)(y) \wedge f(Q_s^x)(y) = f((R \cap Q)_s^x)(y)$ for all $y \in V$. In fact,

$$
\begin{aligned}
f(R_s^x)(y) \wedge f(Q_s^x)(y) &= [\vee\{R_s^x(z) \mid z \in U \text{ and } f(z) = y\}] \wedge [\vee\{Q_s^x(z) \mid z \in U \text{ and } f(z) = y\}] \\
&= r_y \wedge [\vee\{Q_s^x(z) \mid z \in U \text{ and } f(z) = y\}] \\
&= \vee\{r_y \wedge Q_s^x(z) \mid z \in U \text{ and } f(z) = y\} \\
&= \vee\{R_s^x(z) \wedge Q_s^x(z) \mid z \in U \text{ and } f(z) = y\} \\
&= \vee\{(R_s^x \cap Q_s^x)(z) \mid z \in U \text{ and } f(z) = y\} \\
&= \vee\{(R \cap Q)_s^x(z) \mid z \in U \text{ and } f(z) = y\} \\
&= f((R \cap Q)_s^x)(y),
\end{aligned}
$$

i.e., $f(R_s^x)(y) \wedge f(Q_s^x)(y) = f((R \cap Q)_s^x)(y)$, as desired. Hence, the first assertion holds.

(2) Again, without loss of generality, we may assume that $f$ is successor-consistent with respect to $R$. Whence, $f$ is predecessor-consistent with respect to $R^{-1}$ by Proposition 2.1. It follows from the first assertion and Eqs. (1), (2), (5), and (6) that

$$
\begin{aligned}
f((R \cap Q)_p^x) &= f(R_p^x \cap Q_p^x) \\
&= f((R^{-1})_s^x \cap (Q^{-1})_s^x) \\
&= f((R^{-1} \cap Q^{-1})_s^x) \\
&= f((R^{-1})_s^x) \cap f((Q^{-1})_s^x) \\
&= f(R_p^x) \cap f(Q_p^x),
\end{aligned}
$$

namely, $f((R \cap Q)_p^x) = f(R_p^x) \cap f(Q_p^x)$, finishing the proof of the theorem. □

For the union operation, any mapping preserves fuzzy predecessor neighborhoods and fuzzy successor neighborhoods.

**Proposition 2.2.** *Let R and Q be fuzzy relations on U, and $f : U \longrightarrow V$ a mapping. Then for any $x \in U$,*

(1) $f((R \cup Q)_p^x) = f(R_p^x) \cup f(Q_p^x)$.
(2) $f((R \cup Q)_s^x) = f(R_s^x) \cup f(Q_s^x)$.

*Proof.* It follows directly from Eqs. (3) and (4). □

The next theorem presents an equivalent characterization of predecessor-consistent (successor-consistent) functions.

**Theorem 2.4.** *Let R be a fuzzy relation on U, and $f : U \longrightarrow V$ a mapping.*

(1) *The mapping $f$ is predecessor-consistent with respect to R if and only if $f^{-1}(f(R_s^x)) = R_s^x$ for any $x \in U$.*
(2) *The mapping $f$ is successor-consistent with respect to R if and only if $f^{-1}(f(R_p^x)) = R_p^x$ for any $x \in U$.*

*Proof.* (1) For the 'if' part, suppose, by contradiction, that there are $x_1, x_2 \in U$ with $f(x_1) = f(x_2)$ such that $R_p^{x_1} \neq R_p^{x_2}$. Without loss of generality, assume that there exists some $z \in U$ such that $R_p^{x_1}(z) > R_p^{x_2}(z)$. On the other hand, we have by condition that $f^{-1}(f(R_s^z)) = R_s^z$. It follows that

$$
\begin{aligned}
R_s^z(x_2) &= f^{-1}(f(R_s^z))(x_2) \\
&= f(R_s^z)(f(x_2)) \\
&= \vee\{R_s^z(x) \mid x \in U \text{ and } f(x) = f(x_2)\} \\
&\geq R_s^z(x_1),
\end{aligned}
$$

namely, $R_s^z(x_2) \geq R_s^z(x_1)$. Clearly, it is equivalent to that $R_p^{x_2}(z) \geq R_p^{x_1}(z)$. This, together with the assumption $R_p^{x_1}(z) > R_p^{x_2}(z)$, forces that $R_p^{x_2}(z) > R_p^{x_2}(z)$, which is absurd. Therefore, $R_p^{x_1} = R_p^{x_2}$, and the sufficiency holds.

To see the 'only if' part, suppose that $f$ is predecessor-consistent with respect to $R$. As we claimed in the proof of Theorem 2.3, if $y, z \in U$ with $f(y) = f(z)$, then $R_s^x(y) = R_s^x(z)$ for any $x \in U$. Consequently, we obtain that for any $x \in U$,

$$
\begin{aligned}
f^{-1}(f(R_s^x))(z) &= f(R_s^x)(f(z)) \\
&= \vee\{R_s^x(y) \mid y \in U \text{ and } f(y) = f(z)\} \\
&= R_s^x(z),
\end{aligned}
$$

for all $z \in U$, i.e., $f^{-1}(f(R_s^x)) = R_s^x(z)$, as desired. Hence, the first assertion holds.

(2) By Proposition 2.1, $f$ is successor-consistent with respect to $R$ if and only if it is predecessor-consistent with respect to $R^{-1}$. By the first assertion, this is equivalent to $f^{-1}(f((R^{-1})_s^x)) = (R^{-1})_s^x$ for any $x \in U$. Further, this is equivalent to $f^{-1}(f(R_p^x)) = R_p^x$ for any $x \in U$, as $(R^{-1})_s^x = R_p^x$. Thereby, the assertion (2) is true and this finishes the proof of the theorem. □

## 3. Fuzzy relation mappings

In order to develop tools for studying the communication between two fuzzy information systems, [12] explored fuzzy relation mappings and their properties. This section is devoted to extending and improving these properties.

Let us review the definition of fuzzy relation mappings obtained by Zadeh's extension principle.

**Definition 3.1.** Let $U$ and $V$ be nonempty universal sets, and $f : U \longrightarrow V$ a mapping.

(1) The *fuzzy relation mapping* induced by $f$, denoted by the same notation $f$, is a mapping from $\mathcal{F}(U \times U)$ to $\mathcal{F}(V \times V)$ that maps $R$ to $f(R)$, where $f(R)$ is defined by

$$f(R)(y_1, y_2) = \vee\{R(x_1, x_2) \mid x_i \in U, f(x_i) = y_i, i = 1, 2\}$$

for all $(y_1, y_2) \in V \times V$.

(2) The *inverse fuzzy relation mapping* induced by $f$, denoted by $f^{-1}$, is a mapping from $\mathcal{F}(V \times V)$ to $\mathcal{F}(U \times U)$ that maps $Q$ to $f^{-1}(Q)$, where $f^{-1}(Q)$ is defined by

$$f^{-1}(Q)(x_1, x_2) = Q(f(x_1), f(x_2))$$

for all $(x_1, x_2) \in U \times U$.

To illustrate the above definition, let us revisit Example 2.1.

**Example 3.1.** Recall that in Example 2.1, $U = \{x_1, x_2, \ldots, x_8\}$, $V = \{y_1, y_2, \ldots, y_8\}$, and

$$R = \frac{1}{(x_1, x_2)} + \frac{1}{(x_1, x_3)} + \frac{0.8}{(x_2, x_4)} + \frac{0.8}{(x_2, x_5)} + \frac{0.9}{(x_3, x_4)} + \frac{0.8}{(x_3, x_5)} + \frac{0.7}{(x_4, x_6)} + \frac{0.7}{(x_4, x_7)} +$$
$$\frac{0.7}{(x_5, x_6)} + \frac{0.7}{(x_5, x_7)} + \frac{0.9}{(x_6, x_8)} + \frac{0.9}{(x_7, x_8)}.$$

Consider $f_1 : U \longrightarrow V$ defined by

$$f_1(x_i) = \begin{cases} y_2 & \text{if } i = 2, 3, \\ y_i & \text{otherwise.} \end{cases}$$

Then it follows by definition that

$$f_1(R) = \frac{1}{(y_1, y_2)} + \frac{0.9}{(y_2, y_4)} + \frac{0.8}{(y_2, y_5)} + \frac{0.7}{(y_4, y_6)} + \frac{0.7}{(y_4, y_7)} + \frac{0.7}{(y_5, y_6)} + \frac{0.7}{(y_5, y_7)} + \frac{0.9}{(y_6, y_8)} + \frac{0.9}{(y_7, y_8)},$$
$$f_1^{-1}(f_1(R)) = \frac{1}{(x_1, x_2)} + \frac{1}{(x_1, x_3)} + \frac{0.9}{(x_2, x_4)} + \frac{0.8}{(x_2, x_5)} + \frac{0.9}{(x_3, x_4)} + \frac{0.8}{(x_3, x_5)} + \frac{0.7}{(x_4, x_6)} + \frac{0.7}{(x_4, x_7)} +$$
$$\frac{0.7}{(x_5, x_6)} + \frac{0.7}{(x_5, x_7)} + \frac{0.9}{(x_6, x_8)} + \frac{0.9}{(x_7, x_8)}.$$

Recall that in [12], Theorem 2.4(4) says that the transitivity of $R$ implies that of $f(R)$ when the mapping $f : U \longrightarrow V$ is surjective and consistent (i.e., both predecessor-consistent and successor-consistent) with respect to $R \in \mathcal{F}(U \times U)$. In fact, the requirement that $f$ is surjective is not necessary; and moreover, either of predecessor-consistency or successor-consistency is enough.

**Theorem 3.1.** *Let $U$ and $V$ be finite universal sets. Suppose that $f : U \longrightarrow V$ is a mapping and $R \in \mathcal{F}(U \times U)$ is transitive. Then $f(R)$ is transitive if one of the following conditions holds:*

(1) *$f$ is predecessor-consistent with respect to $R$.*
(2) *$f$ is successor-consistent with respect to $R$.*

*Proof.* For $f(R)$ to be transitive, we must show that $f(R)(y_1, y_3) \geq f(R)(y_1, y_2) \wedge f(R)(y_2, y_3)$ for any $y_1, y_2, y_3 \in V$. For simplicity, we write $r_1, r_2, r_3$ for $f(R)(y_1, y_3)$, $f(R)(y_1, y_2)$, and $f(R)(y_2, y_3)$, respectively. Hence, we need to verify that $r_1 \geq r_2 \wedge r_3$. Note that $f(R)(y_1, y_3) = \vee \{R(x_1, x_3) \mid x_i \in U, f(x_i) = y_i, i = 1, 3\}$ by definition. Therefore, there are $a_1, a_3 \in U$ with $f(a_1) = y_1$ and $f(a_3) = y_3$ such that $R(a_1, a_3) = r_1$. Similarly, there are $b_1, b_2 \in U$ with $f(b_1) = y_1$ and $f(b_2) = y_2$ such that $R(b_1, b_2) = r_2$, and there are $c_2, c_3 \in U$ with $f(c_2) = y_2$ and $f(c_3) = y_3$ such that $R(c_2, c_3) = r_3$.

For (1), assume that $f$ is predecessor-consistent with respect to $R$. As $f(b_2) = f(c_2)$, we get by the definition of predecessor-consistent functions that $R_p^{b_2} = R_p^{c_2}$, which means that $R(b_1, c_2) = R_p^{c_2}(b_1) = R_p^{b_2}(b_1) = R(b_1, b_2) = r_2$, i.e., $R(b_1, c_2) = r_2$. This, together with $R(c_2, c_3) = r_3$, gives rise to $R(b_1, c_3) \geq r_2 \wedge r_3$ since $R$ is transitive. On the other hand, we have that $R(b_1, c_3) \leq \vee \{R(x_1, x_3) \mid x_i \in U, f(x_i) = y_i, i = 1, 3\} = f(R)(y_1, y_3) = r_1$, namely, $R(b_1, c_3) \leq r_1$. Whence, $r_1 \geq r_2 \wedge r_3$, as desired.

For (2), assume that $f$ is successor-consistent. It follows from $f(b_2) = f(c_2)$ that $R_s^{b_2} = R_s^{c_2}$, which means that $R(b_2, c_3) = R_s^{b_2}(c_3) = R_s^{c_2}(c_3) = R(c_2, c_3) = r_3$, i.e., $R(b_2, c_3) = r_3$. This, together with $R(b_1, b_2) = r_2$, gives $R(b_1, c_3) \geq r_2 \wedge r_3$ since $R$ is transitive. It forces by the previous argument $R(b_1, c_3) \leq r_1$ that $r_1 \geq r_2 \wedge r_3$. Therefore, $f(R)$ is transitive, finishing the proof of the theorem. □

Let $f : U \longrightarrow V$ be a mapping, and $R, Q \in \mathcal{F}(U \times U)$. In [12], Theorem 2.5(2) says that $f(R \cap Q) = f(R) \cap f(Q)$ if $f$ is consistent (i.e., both predecessor-consistent and successor-consistent) with respect to both $R$ and $Q$. We now show that the requirement of $f$ can be relaxed as follows.

**Theorem 3.2.** *Let $U$ and $V$ be finite universal sets, $f : U \longrightarrow V$ a mapping, and $R, Q \in \mathcal{F}(U \times U)$. Then $f(R \cap Q) = f(R) \cap f(Q)$ if one of the following conditions holds.*

(1) *The mapping $f$ is both predecessor-consistent and successor-consistent with respect to $R$.*
(2) *The mapping $f$ is both predecessor-consistent and successor-consistent with respect to $Q$.*

(3) *The mapping $f$ is predecessor-consistent with respect to $R$ and successor-consistent with respect to $Q$.*

(4) *The mapping $f$ is successor-consistent with respect to $R$ and predecessor-consistent with respect to $Q$.*

*Proof.* We only prove (1) and (3), because of the symmetry of the assertions. Let us begin with (1). Since $f$ is both predecessor-consistent and successor-consistent with respect to $R$, we have by Theorem 2.1 that $R(x_1, x_2) = R(x'_1, x'_2)$ for any $x_i, x'_i \in U$ satisfying $f(x_i) = f(x'_i)$, where $i = 1, 2$. In light if this, we may write $r$ for all $R(x_1, x_2)$ with $f(x_1) = y_1$ and $f(x_2) = y_2$. In fact, $r$ only depends on $y_1$ and $y_2$. It thus follows that

$$
\begin{aligned}
(f(R) \cap f(Q))(y_1, y_2) &= f(R)(y_1, y_2) \wedge f(Q)(y_1, y_2) \\
&= [\vee \{R(x_1, x_2) \mid x_i \in U, f(x_i) = y_i, i = 1, 2\}] \wedge [\vee \{Q(x_1, x_2) \mid x_i \in U, f(x_i) = y_i, i = 1, 2\}] \\
&= r \wedge [\vee \{Q(x_1, x_2) \mid x_i \in U, f(x_i) = y_i, i = 1, 2\}] \\
&= \vee \{r \wedge Q(x_1, x_2) \mid x_i \in U, f(x_i) = y_i, i = 1, 2\} \\
&= \vee \{R(x_1, x_2) \wedge Q(x_1, x_2) \mid x_i \in U, f(x_i) = y_i, i = 1, 2\} \\
&= \vee \{(R \cap Q)(x_1, x_2) \mid x_i \in U, f(x_i) = y_i, i = 1, 2\} \\
&= f(R \cap Q)(y_1, y_2),
\end{aligned}
$$

for any $y_1, y_2 \in U$. Hence, $f(R \cap Q) = f(R) \cap f(Q)$ in this case.

For (3), note that $f(R \cap Q) \subseteq f(R) \cap f(Q)$ always holds by definition. Hence, we need only to verify the inverse inclusion, that is, $f(R \cap Q)(y_1, y_2) \geq f(R)(y_1, y_2) \wedge f(Q)(y_1, y_2)$ for all $y_1, y_2 \in U$. Because $f$ is predecessor-consistent with respect to $R$, we obtain that

$$
\begin{aligned}
f(R)(y_1, y_2) &= \vee \{R(x_1, x_2) \mid x_i \in U, f(x_i) = y_i, i = 1, 2\} \\
&= \bigvee_{x_1 \in f^{-1}(y_1)} \bigvee_{x_2 \in f^{-1}(y_2)} R(x_1, x_2) \\
&= \bigvee_{x_1 \in f^{-1}(y_1)} R(x_1, b),
\end{aligned}
$$

where $b \in f^{-1}(y_2)$. Clearly, there is $a \in f^{-1}(y_1)$ such that $R(a, b) = f(R)(y_1, y_2)$. On the other hand, since $f$ is successor-consistent with respect to $Q$, we have that

$$
\begin{aligned}
f(Q)(y_1, y_2) &= \vee \{Q(x_1, x_2) \mid x_i \in U, f(x_i) = y_i, i = 1, 2\} \\
&= \bigvee_{x_2 \in f^{-1}(y_2)} \bigvee_{x_1 \in f^{-1}(y_1)} Q(x_1, x_2) \\
&= \bigvee_{x_2 \in f^{-1}(y_2)} Q(a', x_2),
\end{aligned}
$$

where $a' \in f^{-1}(y_1)$. Clearly, there exists $b' \in f^{-1}(y_2)$ such that $Q(a', b') = f(Q)(y_1, y_2)$. Furthermore, we get by the consistency of $f$ that

$$
\begin{aligned}
f(R \cap Q)(y_1, y_2) &= \vee \{(R \cap Q)(x_1, x_2) \mid x_i \in U, f(x_i) = y_i, i = 1, 2\} \\
&= \vee \{R(x_1, x_2) \wedge Q(x_1, x_2) \mid x_i \in U, f(x_i) = y_i, i = 1, 2\} \\
&= \bigvee_{x_1 \in f^{-1}(y_1)} \bigvee_{x_2 \in f^{-1}(y_2)} [R(x_1, x_2) \wedge Q(x_1, x_2)] \\
&\geq \bigvee_{x_1 \in f^{-1}(y_1)} [R(x_1, b') \wedge Q(x_1, b')] \\
&= \bigvee_{x_1 \in f^{-1}(y_1)} [R(x_1, b) \wedge Q(x_1, b')] \\
&\geq R(a, b) \wedge Q(a, b') \\
&= R(a, b) \wedge Q(a', b') \\
&= f(R)(y_1, y_2) \wedge f(Q)(y_1, y_2).
\end{aligned}
$$

That is, $f(R \cap Q)(y_1, y_2) \geq f(R)(y_1, y_2) \wedge f(Q)(y_1, y_2)$ for any $y_1, y_2 \in U$. Consequently, $f(R \cap Q) = f(R) \cap f(Q)$ in the case of (3). This completes the proof of the theorem. □

The next theorem extends the assertion (2) of Theorem 2.7 in [12], where only the sufficiency has been provided.

**Theorem 3.3.** *Let $f : U \longrightarrow V$ be a mapping and $R \in \mathcal{F}(U \times U)$. Then $f^{-1}(f(R)) = R$ if and only if $f$ is both predecessor-consistent and successor-consistent with respect to $R$.*

*Proof.* We only verify the necessity here; the reader may refer to [12] for the proof of the sufficiency. Assume, by contradiction, that $f$ is not predecessor-consistent. Then there are $x_1, x_2 \in U$ with $f(x_1) = f(x_2)$ such that $R_p^{x_1} \neq R_p^{x_2}$. Thereby, there is some $z \in U$ such that $R_p^{x_1}(z) \neq R_p^{x_2}(z)$, namely, $R(z, x_1) \neq R(z, x_2)$. It follows from $f^{-1}(f(R)) = R$ that $f^{-1}(f(R))(z, x_1) \neq f^{-1}(f(R))(z, x_2)$. We get by definition that $f(R)(f(z), f(x_1)) \neq f(R)(f(z), f(x_2))$. It is a contradiction as $f(x_1) = f(x_2)$. As a result, $f$ is predecessor-consistent with respect to $R$. Similarly, it is easy to show that $f$ is also successor-consistent with respect to $R$. Therefore, the necessity holds. $\square$

Let us end this section with a relationship between fuzzy neighborhoods and fuzzy relation mappings, which provides an approach to computing the fuzzy predecessor and successor neighborhoods of an element of $V$ with respect to $f(R)$.

**Theorem 3.4.** *Let $f : U \longrightarrow V$ be a mapping and $R \in \mathcal{F}(U \times U)$. Then for any $y \in V$,*

(1) $f(R)_p^y = \bigcup_{x \in f^{-1}(y)} f(R_p^x)$. *In particular, $f(R)_p^y = f(R_p^x)$ for any $x \in f^{-1}(y)$ if $f$ is predecessor-consistent with respect to $R$.*

(2) $f(R)_s^y = \bigcup_{x \in f^{-1}(y)} f(R_s^x)$. *In particular, $f(R)_s^y = f(R_s^x)$ for any $x \in f^{-1}(y)$ if $f$ is successor-consistent with respect to $R$.*

*Proof.* We only prove the first assertion, since the second one can be proved similarly. Note that if $y \notin f(U)$, then it is clear that $f(R)_p^y = f(R)_s^y = \emptyset$ and the assertion holds. Otherwise, we have that

$$
\begin{aligned}
\left(\bigcup_{x \in f^{-1}(y)} f(R_p^x)\right)(z) &= \bigvee_{x \in f^{-1}(y)} f(R_p^x)(z) \\
&= \bigvee_{x \in f^{-1}(y)} \bigvee_{x' \in f^{-1}(z)} R(x', x) \\
&= \vee\{R(x', x) \mid x, x' \in U, f(x') = z, f(x) = y\} \\
&= f(R)(z, y) \\
&= f(R)_p^y(z),
\end{aligned}
$$

for all $z \in V$. Hence, $f(R)_p^y = \bigcup_{x \in f^{-1}(y)} f(R_p^x)$, as desired.

For any given $x \in f^{-1}(y)$, if $f$ is predecessor-consistent with respect to $R$, then for any $x' \in f^{-1}(y)$, we have by definition that $R_p^{x'} = R_p^x$. This gives rise to $f(R)_p^y = \bigcup_{x' \in f^{-1}(y)} f(R_p^{x'}) = f(R_p^x)$, completing the proof of the first assertion. $\square$

## 4. Conclusion

In this paper, we have introduced predecessor-consistent and successor-consistent functions with respect to a fuzzy relation. They are together equivalent to the notion of consistent functions in the sense of [12]. Some properties of predecessor-consistent and successor-consistent functions have been explored. Based on the classification of consistent functions, we have greatly improved some characterizations of fuzzy relation mappings presented in [12]. The results obtained in the paper can help us establish a homomorphism between two fuzzy information systems and further compare their properties.

## Acknowledgements

This work was supported by the National Natural Science Foundation of China under Grants 60821001, 60873191, and 60903152.